\documentclass{svproc}
\usepackage{url}

\usepackage{amsfonts}
\usepackage[margin=1.2in]{geometry}
\usepackage{graphicx}
\usepackage{multirow}
\usepackage{float}
\usepackage{lastpage}
\usepackage{fancyhdr}

\usepackage{colortbl}
\usepackage{color}
\usepackage{array, makecell}
\usepackage{boldline}
\usepackage[x11names, table]{xcolor}

\newcolumntype{?}[1]{!{\vrule width #1}}
\definecolor{colT1}{rgb}{0.95,0.88,0.2}
\definecolor{colT2}{rgb}{0.98,0.95,0.8}
\definecolor{colT3}{rgb}{0.0,0.0,0.0}
\definecolor{colT4}{rgb}{1.0,1.0,1.0}

\begin{document}
\mainmatter              
\title{Data Augmentation and Clustering for Vehicle Make/Model Classification}
\titlerunning{Vehicle Make/Model Classification}  
%
\author{Mohamed Nafzi \and Michael Brauckmann\inst{1} \and Tobias Glasmachers\inst{2}}
\authorrunning{Mohamed Nafzi et al.} 
%
\tocauthor{Ivar Ekeland, Roger Temam, Jeffrey Dean, David Grove,
Craig Chambers, Kim B. Bruce, and Elisa Bertino}
\institute{Facial \& Video Analytics\\
	IDEMIA Identity \& Security Germany AG\\
\email{mohamed.nafzi@idemia.com, michael.brauckmann@idemia.com},\\
\and
Institute for Neural Computation\\
Ruhr-University Bochum, Germany\\
\email{tobias.glasmachers@ini.rub.de}}

\maketitle              

\begin{abstract}
Vehicle shape information is very important in Intelligent Traffic Systems (ITS). In this paper we present a way to exploit a training data set of vehicles released in different years and captured under different perspectives. Also the efficacy of clustering to enhance the make/model classification is presented. Both steps led to improved classification results and a greater robustness. Deeper convolutional neural network based on ResNet architecture has been designed for the training of the vehicle make/model classification. The unequal class distribution of training data produces an a priori probability. Its elimination, obtained by removing of the bias and through hard normalization of the centroids in the classification layer, improves the classification results. A developed application has been used to test the vehicle re-identification on video data manually based on make/model and color classification. This work was partially funded under the grant. 
\end{abstract}

\section{Introduction}

The aim of this work is the improvement of the vehicle re-identification module based on shape and color classification from our previous work \cite{paper18} but also to present methods, which definitely help other researcher to improve the accuracy of their vehicle classification module. To reach this goal, we had to overcome several challenges. The appearance of a vehicle varies not only due to its make and to its model but also can differ strongly depending on the year of released and the perspective. For this reason, we created a data augmentation process employing a developed web crawler querying vehicles with different model and their years and different views, such as front, rear or side. The training data labels contain only make and model information, no year of release or view information. To obtain a refined underlying representation for the two missing labels, a clustering approach was developed. It shows a hierarchical structure, to generate data driven different subcategories for each make and model label pair. In this work, we used convolutional neural network (CNN) to train a vehicle make/model classifier based on the ResNet architecture. We performed a threshold optimization for the make/model and color classification to suppress false classifications and detections. Comparing with the works of other researchers, our module contains more classes to classify. It covers most of the known makes and models of the released years between 1990 and 2018 worldwide.

\section{Related Works}
Some research has been performed on make/model classification of vehicles. Most of it operated on a small number of make/models because it is difficult to get a labeled data set panning all existing make/models. Manual annotation is almost impossible because one needs an expert for each make being able to recognize all its models and it is very tedious and time consuming process. \cite{paper12} presented a vehicle make/model classification trained on just 5 classes. 3 different classifiers were tested based on one class k-nearest-neighbor, multi-class and neural network and operated on frontal images. \cite{paper8} used 3D curve alignment and trained just 6 make/models. \cite{paper16} developed a classifier based on a probabilistic neural network using 10 different classes. They worked on frontal images. \cite{paper6} applied image segmentation and combined local and global descriptors to classify 10 classes. \cite{paper4} used a classifier based on CNN to classify 13 different make/models. \cite{paper9} used symmetrical speeded up robust features to separate between 29 classes. \cite{paper15} showed a classification based on 3D-model using 36 classes in a Bayesian approach supporting multiple views. \cite{paper7} presented a make/model identification of 50 make/models based on oriented contour points. Two strategies have been tested, the discriminant function combining several classification scores and voting spaces. \cite{paper11} used the geometry and the appearance of car emblems from rear view to identify 52 make/models. Different features and classifications have been tested. \cite{paper13} investigated two different classification approaches, a k-nearest-neighbour classifier and a naive bayes classifier and worked on 74 different classes with frontal images. \cite{paper14} developed a make/model classification based on feature representation for rigid structure recognition using 77 different classes. Two distances have been tested, the dot product and the euclidean distance. \cite{paper5} tested different methods by make/model classification of 86 different classes on images with side view. The best one was HoG-RBF-SVM. \cite{paper17} used 3D-boxes of the image with its rasterized low-resolution shape and information about the 3D vehicle orientation as CNN-input to classify 126 different make/models. The module of \cite{paper10} is based on 3D object representations using linear SVM classifiers and trained on 196 classes. In a real video scene all existing make/models could occur. Considering that we have worldwide more than 2000 models, make/model classification trained just on few classes will not succeed in practical applications. \cite{paper1} increase the number of the trained classes. His module is based on CNN and  trained on 59 different vehicle makes as well as on 818 different models. His solution seems to be closer for commercial use. Our developed module in our previous work \cite{paper18} was trained on 1447 different classes and could recognize 137 different vehicle makes as well as 1447 different models of the released year between 2016 till 2018. In this paper We made a data augmentation and clustering. Our current module was trained on 4730 different classes and could recognize 137 different vehicle makes as well as 1447 different models of the released year between 1990 and 2018. It shows better results on video data than \cite{paper18} and \cite{paper1}. We believe that our module gives the appropriate solution for commercial use.

\section{ Data Augmentation}
We used a web crawler to download data with different older model years from 1990 to 2018 and different perspectives. This was an augmentation for our training data. First of all, we removed images with the same content to reduce the redundancy followed by a data cleansing step based on a vehicle quality metric to remove images with bad detections. This data augmentation was an important step to classify more classes.

\section{Clustering}
\begin{figure}[bth]
	\centering
	\begin{minipage}[c]{0.10\textwidth}
		\includegraphics[width=\linewidth]{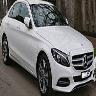}
	\end{minipage}
	\begin{minipage}[c]{0.10\textwidth}
		\includegraphics[width=\linewidth]{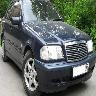}
	\end{minipage}
	\begin{minipage}[c]{0.10\textwidth}		
		\includegraphics[width=\linewidth]{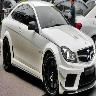}
	\end{minipage}
	\begin{minipage}[c]{0.10\textwidth}
		\includegraphics[width=\linewidth]{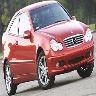}
	\end{minipage}
	\begin{minipage}[c]{0.10\textwidth}
		\includegraphics[width=\linewidth]{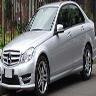}
	\end{minipage}
	\begin{minipage}[c]{0.10\textwidth}
		\includegraphics[width=\linewidth]{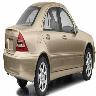}
	\end{minipage}
	\begin{minipage}[c]{0.10\textwidth}		
		\includegraphics[width=\linewidth]{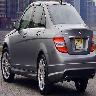}
	\end{minipage}
	\vspace*{2mm}
	\caption{Examples of the generated classes after data augmentation and clustering by Mercedes-Benz C.}
	\label{CS7}
\end{figure}

First, we want to explain why we need clustering. Many vehicles with the same model but different released year or different perspective have less similarity than some vehicles with different model. When we set all elements with the same make/model but different views and different released years in the same class, we get after the training shape feature vectors which are not very close to their centroids. This leads oft to miss classification.
Therefore, clustering is necessary because the labels of the model year and the perspective are not available. We developed a hierarchical clustering method to cluster our data. This method comprises the following processing steps:
\begin{enumerate}
	\item calculation of the distances between all feature vectors within the same make and model category, obtained by the networks described in \cite{paper18}.
	\item Iteratively, determination of the elements that form the maximum feature vector density using a given constant threshold. Creation of a new cluster using these elements.
	\item In the last step: only classes with a minimum number of 20 elements are accepted to avoid label errors.
\end{enumerate}
This data augmentation and clustering have increased the number of our classes from 1447 to 4730. As example by Mercedes-Benz C in our previous work \cite{paper18}, we had just one class. As shown in figure \ref{CS7}. Seven classes with different released years and/or different perspectives were obtained.

\begin{figure}[H]
	\centering
	\includegraphics[width=\linewidth]{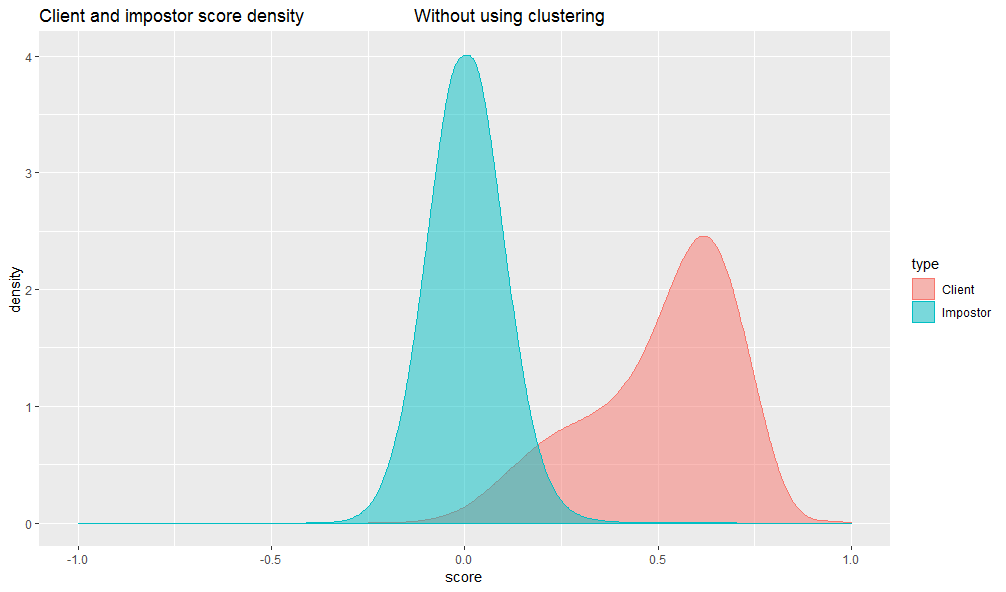}
	\caption{ Density of client and impostor matching scores without applying clustering. \newline \textbf{Client score} is the matching score of two feature vectors with the same model. 
		\newline \textbf{Impostor score} is the matching score of two feature vectors with different model.}
	\label{dcl1}
\end{figure}

\begin{figure}[H]
	\centering
	\includegraphics[width=\linewidth]{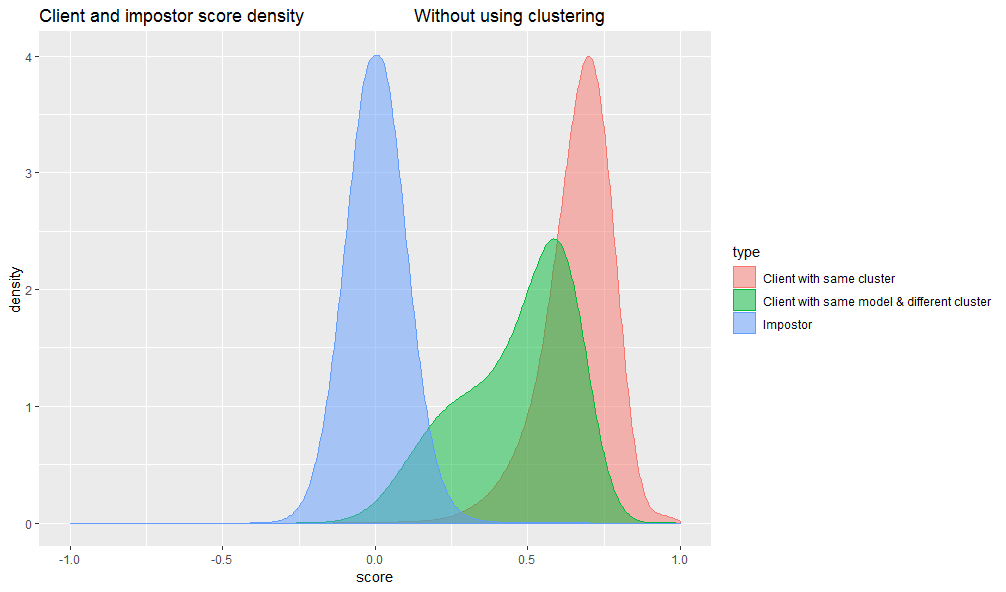}
	\caption{ Density of client and impostor matching scores without applying clustering. Separation of client score density in two densities. \newline \textbf{Client score} is the matching score of two feature vectors with the same model. 
		\newline \textbf{Impostor score} is the matching score of two feature vectors with different model.}
	\label{dcl2}
\end{figure}

\begin{figure}[H]
	\centering
	\includegraphics[width=\linewidth]{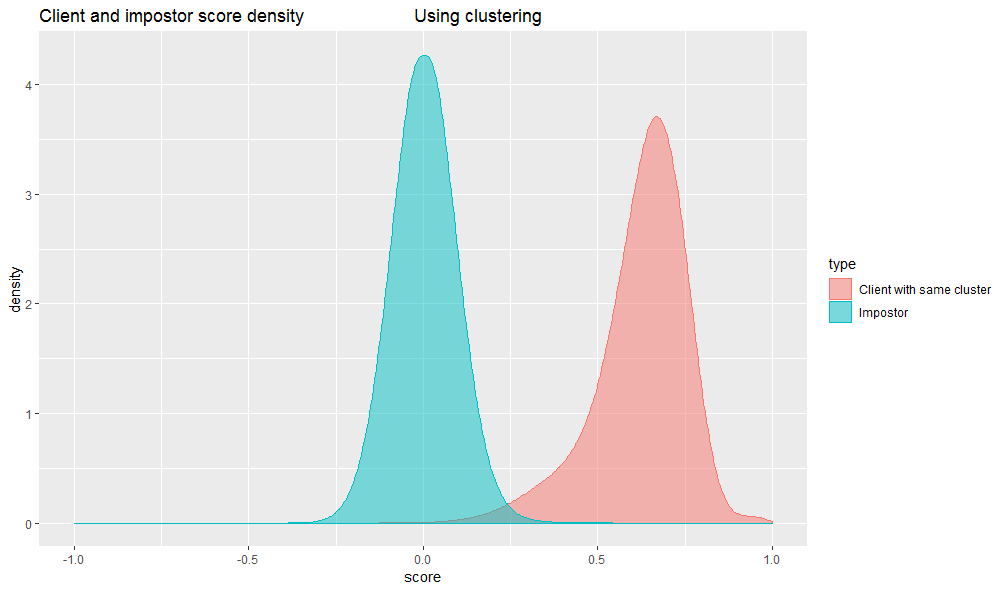}
	\caption{ Density of client and impostor matching scores with applying clustering. \newline \textbf{Client score} is the matching score of two feature vectors with the same cluster. 
		\newline \textbf{Impostor score} is the matching score of two feature vectors with different model.}
	\label{dcl3}
\end{figure}

The figure \ref{dcl1} presents the densities of client and impostor matching scores without applying clustering. Matching of vehicles with the same model but different cluster leads oft to low client scores as shown in the figure \ref{dcl2}. Same model and different cluster means same model and different released years or different views. After applying clustering and retraining the shape network, the feature vectors become very close to their centroids, which decreases the number of low client scores as you could see in the figure \ref{dcl3}. 

\section{CNN-Architectures}
We developed, optimized and tuned our standard CNN-network for make/model classification. It is based on Res-Net architecture. It shows very gut results on controlled data set. Its coding time is 20ms (CPU 1 core, i7-4790, 3.6 GHz). This net was trained on 4730 classes using about 4 Million images.

\section{Removing of A priori Probability}
Our training data does not have equal class distribution. This leads to an a priori probability. The class distribution of testing data differs from training data and is usually unknown. For this reason, we removed this a priori probability by removing of the bias from the classification layer and by normalization of the centroids. To train the modified network faster, we initialized the convolutions and the first IP-layer with our pre-trained network and defined a new classification layer. In this way the training converges after two or three days instead of three or four weeks.

\section{Threshold Optimization}

\begin{figure}[H]
	\centering
	\includegraphics[width=65mm,height=180mm]{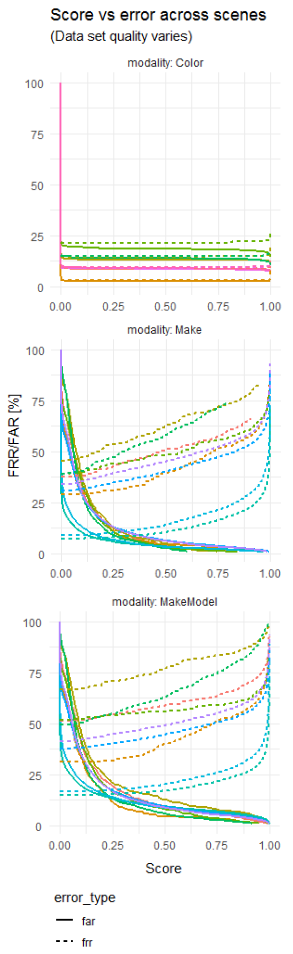}
	\caption{ Error rates of controlled and video data set with different qualities by color, make and make/model classification.\newline This plot helps to set a threshold.\newline FAR: False Acceptance Rate. \newline FRR: False Reject Rate.}
	\label{ts1}
\end{figure}

In practice, it is recommended to set a threshold by the classification. In this way, we could eliminate false classifications. For example, elements with unknown classes, bad image quality, occlusion or false detection leads to miss classification. We calculated the error rates shown in figure \ref{ts1} using different testing data with different qualities to optimize the thresholds for the make/model, make and color classification. We used just similarities with rank 1.

\section{Experiments related to the Clustering of the Training Data and to the Removing of A priori}
We tested two controlled data with good and bad quality. Each data set contains 3306 images with different views and about two images per class. Additionally, we tested Stanford data set and video data. The following steps led to the improvement of our classification module:

\begin{itemize}
	\item Using large scale data set by training.
	\item Cleansing of the data.
	\item Clustering.
	\item Optimization of our CNN-Net.
	\item Removing of a priori.
	\item Using of best-shot by video data.
\end{itemize}

\subsection{Experiments on Controlled Data}

\begin{figure}[H]
	\centering
	\includegraphics[width=\linewidth]{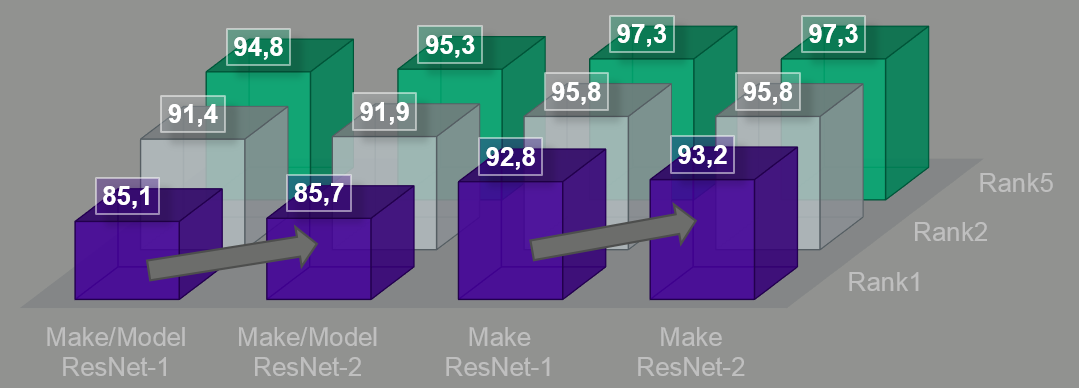}
	\caption{\color{black} Results of make/model classification on controlled data set with good quality and different views. \newline \textbf{ResNet-1} is the trained ResNet in \cite{paper18}. \newline \textbf{ResNet-2} is ResNet-1 + removing of a priori probability.}
	\label{ap1}
\end{figure}

\begin{figure}[H]
	\centering
	\includegraphics[width=\linewidth]{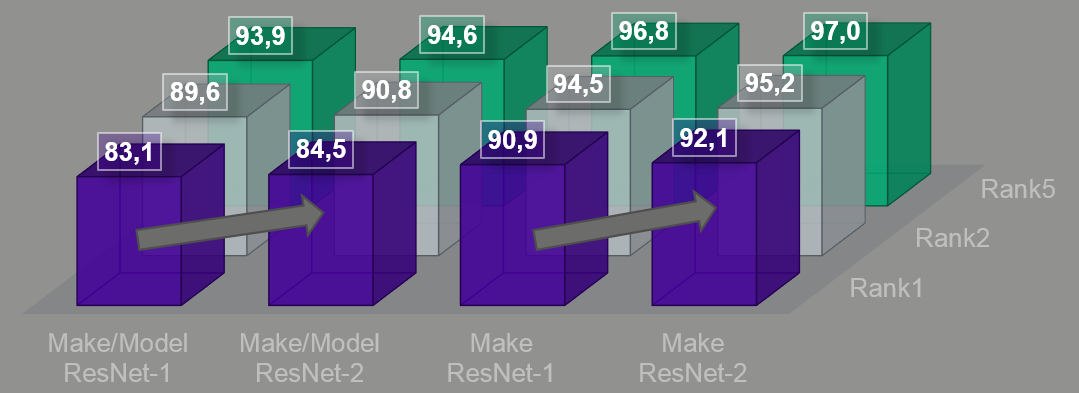}
	\caption{\color{black} Results of make/model classification on controlled data set with bad quality and different views. \newline \textbf{ResNet-1} is the trained ResNet in \cite{paper18}. \newline \textbf{ResNet-2} is ResNet-1 + removing of a priori probability.}
	\label{ap2}
\end{figure}

We tested the effect of removing of a priori probability. The figures \ref{ap1} and \ref{ap2} show the results on controlled data set. The comparison of the results of our make/model classification with the state of the art on Stanford data set has been presented in the table \ref{tb1}.

\color{black}
\color{black}
\begin{table}[htb]
	\caption{Accuracy of our shape module and of the shape module presented by \cite{paper1} on Stanford data set.} 
	\begin{center}
		\begin{tabular}{ ?{0.8pt}c?{0.8pt}c?{0.8pt}c?{0.7pt} }
			\arrayrulecolor{colT3}\hline
			\hlineB{1.5}\cellcolor{colT4} &\cellcolor{colT4} \color{red} Our shape module & \cellcolor{colT4} \cite{paper1} \\ 
			\hline
			\hlineB{1.6}\cellcolor{colT4} Accuracy (top1) & \cellcolor{colT4} \color{red}93.8\% & \cellcolor{colT4} 93.6\% \\
			\hlineB{1.4}\hline
			
		\end{tabular}
	\end{center}
    \label{tb1}
\end{table}

\subsection{Experiments on Video Data}
\begin{figure}[bth]
	\centering
	\includegraphics[width=\linewidth]{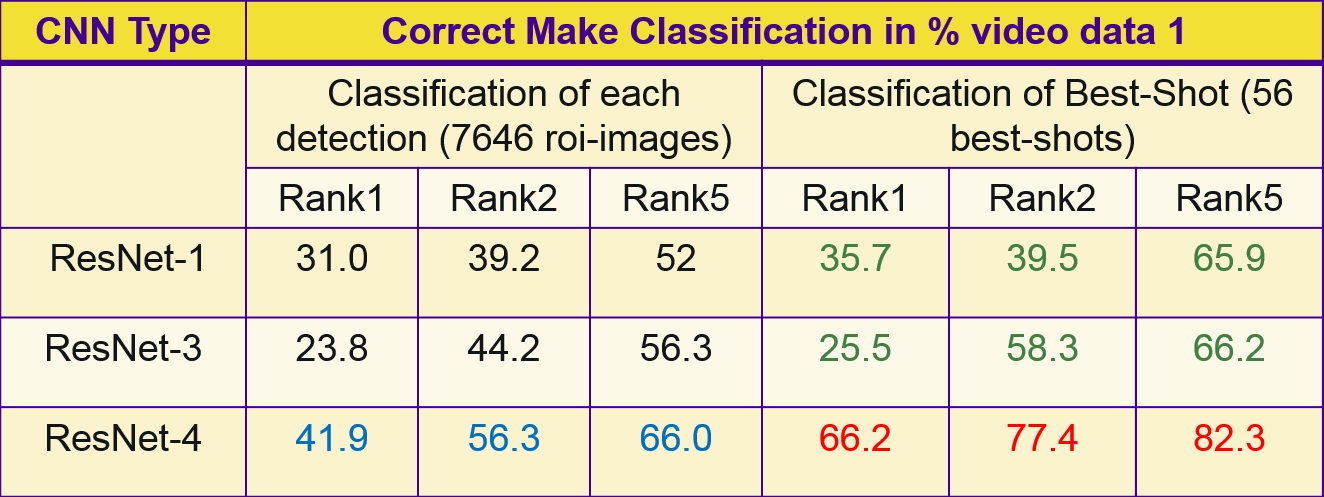}
	\caption{ Results of make classification on traffic video data set with hard quality and different views. 
		\newline \textbf{ResNet-1} is the trained ResNet in \cite{paper18}. 
		\newline \textbf{ResNet-3} is the new trained ResNet using data augmentation (without clustering).
		\newline \textbf{ResNet-4} is the new trained ResNet using data augmentation and clustering.}
	\label{v1}
\end{figure}

\begin{figure}[bth]
	\centering
	\includegraphics[width=\linewidth]{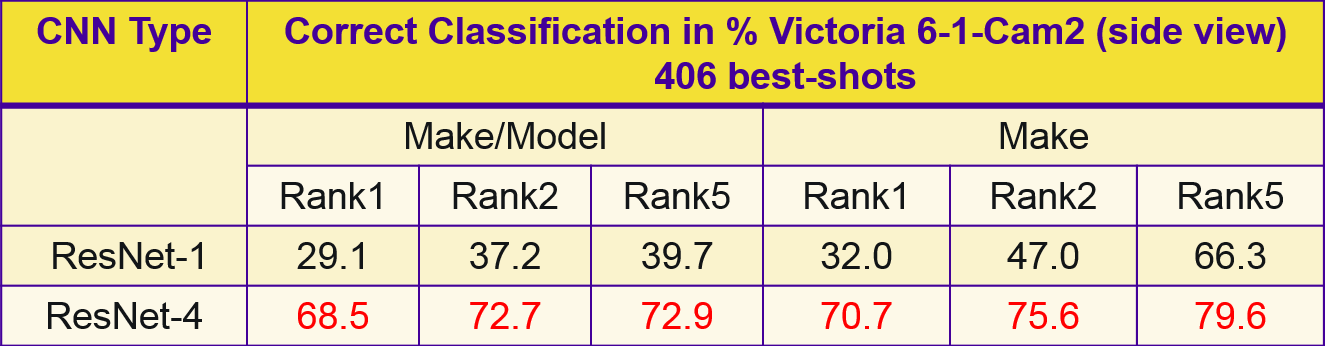}
	\caption{\color{black} Results of make/model and make classification on traffic video data set with medium quality. Sample images shown in figure \ref{t1}.\newline \textbf{ResNet-1} is the trained ResNet in \cite{paper18}.
		\newline \textbf{ResNet-4} is the new trained ResNet using data augmentation and clustering.}
	\label{v2}
\end{figure}

The data augmentation and the clustering leads to better classification than \cite{paper18} especially on video data. Since our video testing data contain some classes, that are not included in our old training data \cite{paper18}. The results in the figure \ref{v1} shows the effect of the data augmentation with and without the use of clustering. Without clustering all elements with the same make/model are in the same class by training. This is not beneficial, because vehicles with the same make/model but different model year and/or different views look different. Therefore, the clustering of the data was necessary. The selection and the classification of the best-Shot ROI-image shows better results than to classify each detection. The tested video data 1 is very challenging because it contains images with some views, which are not included in our training data like top/frontal or top/rear. The results in the figure \ref{v2} shows the effect of the data augmentation using clustering.

\color{black}

\color{black}



%

We used the API referred in \cite{paper1} \color{blue} https://www.sighthound.com/products/cloud \color{black} to test our BCM-Parking1 video data. The API detected correctly the vehicles. The comparison of our results with the results of \cite{paper1} on this video data set shown in the table \ref{tb2} is absolutely fair because we did not tune our training on this data set. It contains 111 best-shots images. 

\begin{table}[H]
	\caption{Accuracy of our shape module and of the shape module presented by \cite{paper1} on our BCM-Parking1 video data. Sample images shown in figure \ref{t2}.
		\newline \textbf{ResNet-1} is the trained ResNet in \cite{paper18}.
		\newline \textbf{ResNet-4} is the new trained ResNet using data augmentation and clustering.} 
	\begin{tabular}{ ?{0.7pt}c?{0.7pt}c?{0.7pt}c?{0.7pt}c?{0.7pt} }
		\arrayrulecolor{colT3}\hline
		\hlineB{1.5}\cellcolor{colT4} &\cellcolor{colT4} ResNet-1 &\cellcolor{colT4} \color{red} ResNet-4 & \cellcolor{colT4} \cite{paper1} \\ 
		\hline
		\hlineB{1.6}\cellcolor{colT4} Make/Model classification (top1) & \cellcolor{colT4} 29.0\% & \cellcolor{colT4} \color{red} 71.1\% & \cellcolor{colT4}  20.5\% \\
		\hline			
		\hlineB{1.6}\cellcolor{colT4} Make classification (top1 )& \cellcolor{colT4} 42.1\% & \cellcolor{colT4} \color{red} 78.9\% & \cellcolor{colT4} 56.4\% \\  
		\hlineB{1.4}\hline
		
	\end{tabular}
	\label{tb2}
\end{table}

\subsection{Tool for Vehicle Re-Identification on Video data}
\begin{figure}[H]
	\centering
	\includegraphics[width=\linewidth]{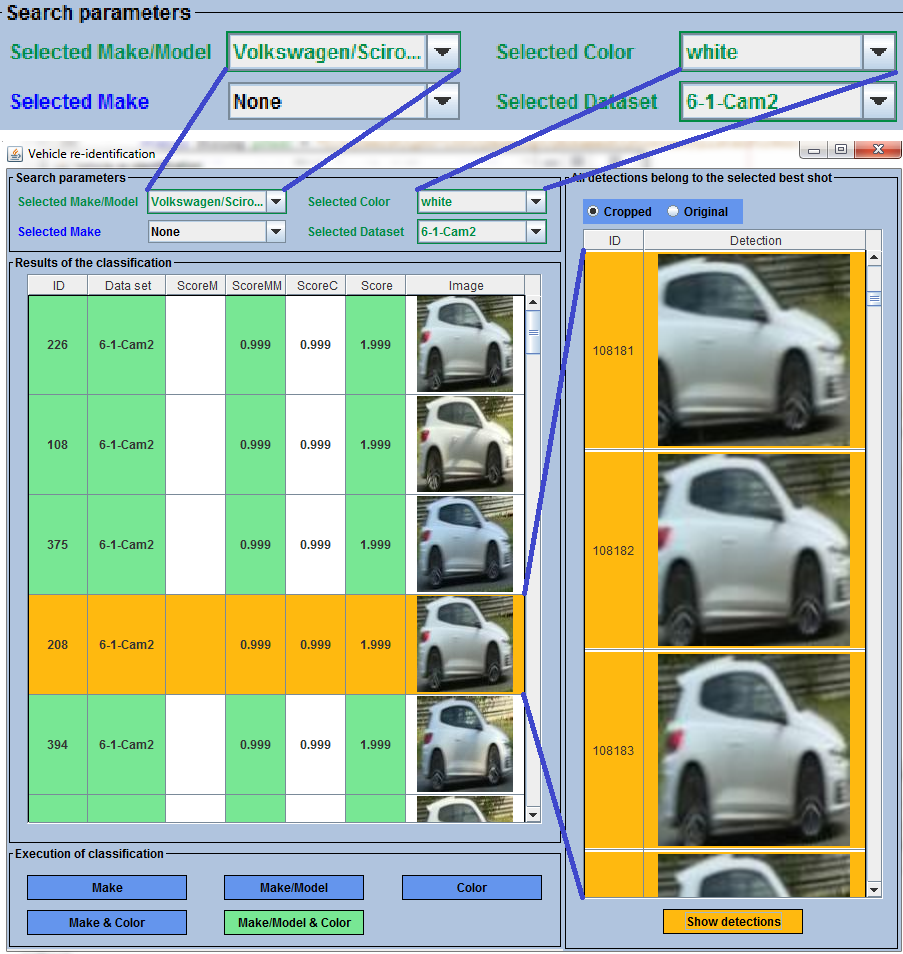}
	\caption{ Vehicle re-identification based on shape (Volkswagen/Scirocco) and color (white) classification (video data).}
	\label{t1}
\end{figure}

\begin{figure}[H]
	\centering
	\includegraphics[width=\linewidth]{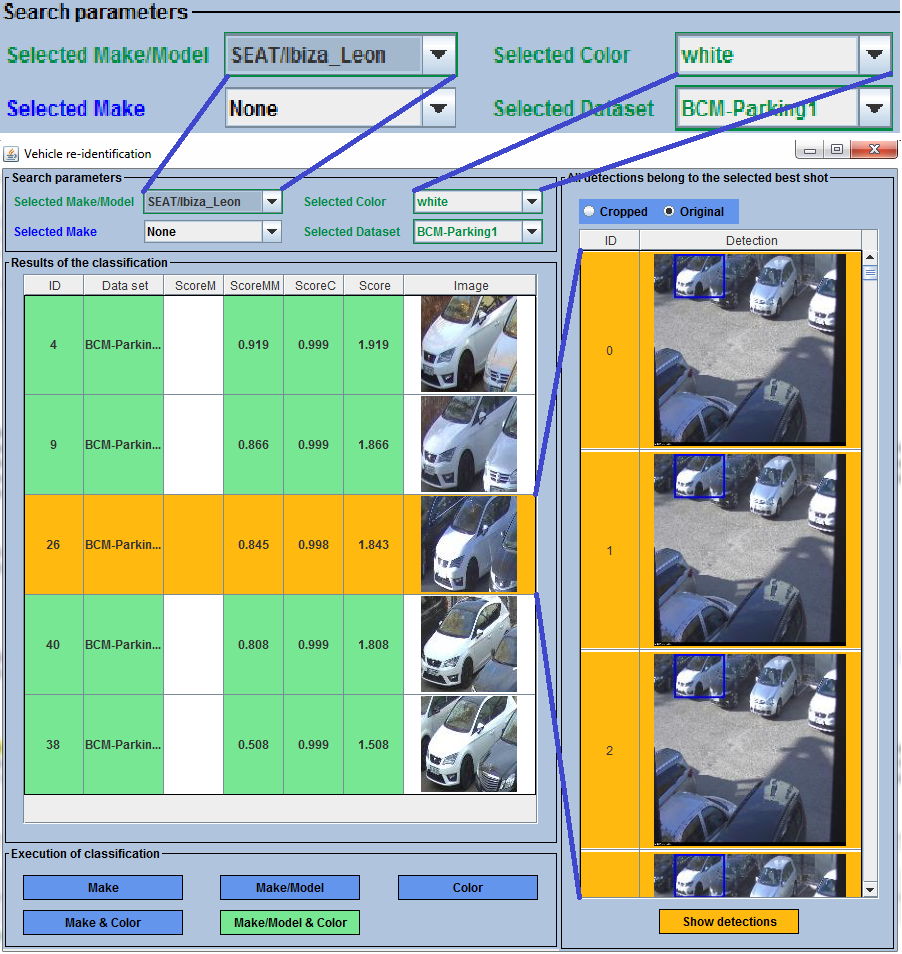}
	\caption{ Vehicle re-identification based on shape (SEAT/Ibiza) and color (white) classification (video data).}
	\label{t2}
\end{figure}

To validate the re-identification module, we developed a stand-alone tool, as shown in figures \ref{t1} and \ref{t2}. This tool allows the re-identification of a vehicle in a video data set with the help of its make or its model. Within the search the combination of shape and color is possible. We used the color classification module of the previous work \cite{paper18}. The results show the best-shot images sorted by their classification scores. We used a threshold to remove false classifications and false detections. The selection of the best-shot image allows to show all ROIs with the same track-id. Further, the original or the ROI-images can be shown. We tested two video data set.  

\section{Conclusion and Future Work}

We presented two methods, that will definitely help other researchers to improve their vehicle shape classification. The first method removing of a priori brings generally 1\% to 2\% absolutely improvement and it is very easy to apply. The second method clustering is necessary and very useful when we make a data augmentation with vehicles with different released years and/or different perspectives and when only the labels make and model are available. In the practice we recommend to set a threshold in term to eliminate bad classifications and bad detections. At the moment we are focussing on the vehicle re-identification based on shape/color feature vector. In this way we could re-identify each vehicle without to know its make/model or its color. Because the features of vehicles with the same shape and color will be close each to other and will produce higher scores by matching if they have similar views. By this method a probe image of the search vehicle is needed by the re-identification.

\section{Acknowledgment}
\begin{itemize}
	\item Victoria: funded by the European Commission (H2020), Grant Agreement number 740754 and is for Video analysis for Investigation of Criminal and Terrorist Activities.
	\item Florida: funded by the German Ministry of Education and Research (BMBF).
\end{itemize}



%
%

\end{document}